\documentclass[letterpaper, 10 pt, conference]{ieeeconf}  

\IEEEoverridecommandlockouts                              
\usepackage{epsfig} 
\usepackage{amsmath} 
\usepackage{amssymb}  

\usepackage{graphicx}
\usepackage{caption}
\usepackage{subcaption}
\usepackage{multicol}
\usepackage{amsmath}
\usepackage{algpseudocode}
\usepackage{algorithm}

\algnewcommand\algorithmicforeach{\textbf{for each}}
\algdef{S}[FOR]{ForEach}[1]{\algorithmicforeach\ #1\ \algorithmicdo}

\title{\LARGE \bf
Coverage Control for a Multi-robot Team with Heterogeneous Capabilities using Block Coordinate Descent (BCD) Method
}

\author{Yung Yu Andy Yiu, Ying Hing Yim, Yan Ning,  Zikai Wang, Ling Shi
\thanks{Yung Yu Andy Yiu, Ying Hing Yim, Yan Ning,  Zikai Wang and Ling Shi are with the Department of Electronic and Computer Engineering, Hong Kong University of Science and Technology, Kowloon, Hong Kong (e-mail: yyayiu@connect.ust.hk; yhyim@connect.ust.hk; yningaa@connect.ust.hk; zwanggz@connect.ust.hk; eesling@ust.hk).}
}

\begin{document}

\setlength{\abovedisplayskip}{10pt}
\setlength{\belowdisplayskip}{10pt}

\newcommand{\norm}[1]{\left\lVert#1\right\rVert_2}

\maketitle
\thispagestyle{empty}
\pagestyle{empty}

\begin{abstract}

In this paper, we propose a coverage control system for a multi-robot team with heterogeneous capabilities to patrol or monitor a bounded environment. The capability could be defined as any criterion of robots like remaining power or mobile speed, depending on the purpose. The proposed control system aims to allocate different portions of the environment to the robots according to their capabilities, i.e., the robot with higher capability takes a larger portion of the environment while the robot with lower capability takes a smaller one. We use the block coordinate descent (BCD) method to optimize the location of portions and the partitioning method alternately. A centralized machine is used to synchronize the robots and the gradient of each robot can be computed in a distributed manner. Simulation results are provided to illustrate the performance of the proposed control system. 

\end{abstract}

\section{INTRODUCTION}
The development of multi-agent systems has grown rapidly in recent years. Compared to single-agent systems, multi-agent systems are more efficient and feasible for tackling some challenging tasks. Research topics on the multi-agent system like formation control, cooperative logistics system, or coverage control are gaining much interest and attention. In this paper, we focus on the coverage control. \par
Coverage control refers to the problem of controlling a group of robots to cover an environment and perform certain tasks, which is significant in many applications. One emergent case is about robots used for monitoring the coastal area and detecting accidents cooperatively. Another case is about robots covering an environment to sense life signals and assist the rescue in a disaster. These applications have pushed the development and conduction of many researches about coverage control from different perspectives.\par
Numbers of research aim to maximize the sensing performance by finding the optimal positions and weights of the robots. For a group of identical robots with sensing cost equals to the square of the distance, Cortés et al.~\cite{Cortes} found that the move-to-centroid algorithm will maximize the overall sensing performance of a known, convex environment. Schwager et al.~\cite{Schwager} extended it to unknown dynamical environments by using sensory information. The algorithm was also extended for non-convex environments by Breitenmoser et al.~\cite{Breitenmoser}. Marier et al.~\cite{Marier} considered the degradation of the sensors on robots, maximized the sensing performance by the move-to-centroid and-compute-weight approach. Carron et al.~\cite{Carron} solved the coverage control problem for unknown sensory functions and provided an algorithm that can perform estimation on sensory functions and coverage control at the same time. \par
The above researches assumed the robots are monitoring their allocated areas from fixed positions. It might be less efficient compared to the approach that considers robots are monitoring their allocated areas by patrolling them, especially when the area is large and the sensing range of each robot is limited. The patrolling approach leads to the consideration of workload balancing in the coverage control problem. Pavone et al.~\cite{Pavone_1}, \cite{Pavone_2} considered the workload balancing tasks with a group of identical robots to cover a non-uniform environment. Pierson et al.~\cite{Pierson_1}, \cite{Pierson_2} first used the move-to-centroid approach to update the robots' positions and then updated the weights by the estimation of actuation and sensing performance. Turanli and Temeltas~\cite{Turanli_1}, \cite{Turanli_2} had a similar approach but proposed a different estimation method. \par
In this paper, we consider the environment coverage task by a team of robots with heterogeneous capabilities. The robots will patrol every point in their allocated cells and hence ratios of the allocated areas are aimed to follow their capabilities ratios. The motivations of our works include: \par
\begin{enumerate}
    \item The capabilities of the robots can be heterogeneous due to many reasons like the difference in robots' types, aging level, etc. They can also change dynamically because of the variation in power consumption. It is more efficient to maintain the assignment of larger cells to the stronger robots.
    \item Most existing methods update the positions of the robots' generators using the move-to-centroid approach proposed by Cortés et al.~\cite{Cortes}, which considers the coverage performance for the robots with fixed positions, instead of patrolling the cells. 
\end{enumerate}
The main contributions are summarized as follows: \par
\begin{enumerate}
    \item We design a coverage control system that optimizes the positions and weights to minimize the error between allocated area ratios and capability ratios, where capabilities can be different initially and can be change dynamically.
    \item We propose a new cost function that directly considers the area ratios. The cost function is differentiable and the gradients can be computed in a distributed manner.
    \item We propose to use the block coordinate descent (BCD) method to optimize the two blocks of variables in the cost function. Gradient based method is used in the optimization for each block. The gradients are also computed in this paper.
\end{enumerate}
The system setup, including the system objective and architecture, will be presented in section \ref{section:setup}. Section \ref{section:method} will introduce the space partitioning methods followed by the optimization including the BCD method and gradient descent method. The proof of convergence will also be provided. Section \ref{section:simulation} shows Matlab simulation results of the proposed coverage control system with changing capabilities and removal of robots. The conclusion will be drawn and some future work will be discussed in section \ref{section:conclusion}.\par

\section{System Setup} \label{section:setup}
In this paper, a group of $n$ robots with positions $\mathcal{X} = \{ x_1, x_2, \ldots, x_n \}$ is considered to cover a known, bounded environment $Q \subset \mathbb{R}^2$ and to perform patrolling or monitoring task. The robots might not be identical and each robot has a non-negative capability factor $C_i \in \mathbb{R}_{\geq 0}$ which reflects its capability to patrol an area. Depending on the purpose, the capability can be defined by any criterion of the robot. For example, it can be a function of its remaining battery power if the target is to balance the duration of the robots, and it can be depended on its mobile speed if we aim to maximize the performance of the system. \par 

\subsection{Objective}
The objective of the system is to allocate different portions of $Q$, denoted as $\mathcal{W} = \{ W_1, W_2, \ldots, W_n \} \subseteq Q^n$, to $n$ different robots according to their capabilities. To generate $\mathcal{W}$, each robot $i$ has its corresponding generator located at $p_i \in Q$ with weight $w_i \in \mathbb{R}$. We define $\mathcal{P}=\{p_1,p_2, \ldots,p_n\}$ as the positions and $\mathbf{w}=\{w_1,w_2,\ldots,w_n\}$ as the weights of all generators. The collection of all $n$ generators is defined as $\mathcal{N} = \{ 1,2, \ldots, n \}$. All the cells $W_i \ \forall i\in \mathcal{N}$ can only overlap on their boundary and they fully cover $Q$, i.e., $\bigcup_{i = 1}^n W_i = Q$. Define the area of $Q$ be $A$ and the area of $W_i$ be $A_i$. The area ratios of the final allocated cells among different robots are aimed to follow the ratios of their capabilities, i.e., $\frac{A_i}{A_j} = \frac{C_i}{C_j} \ , \ \forall i,j \in \mathcal{N}$. \par

\subsection{Systems Architecture}
\begin{figure}[t]
    \centering
    \includegraphics[width=.48\textwidth]{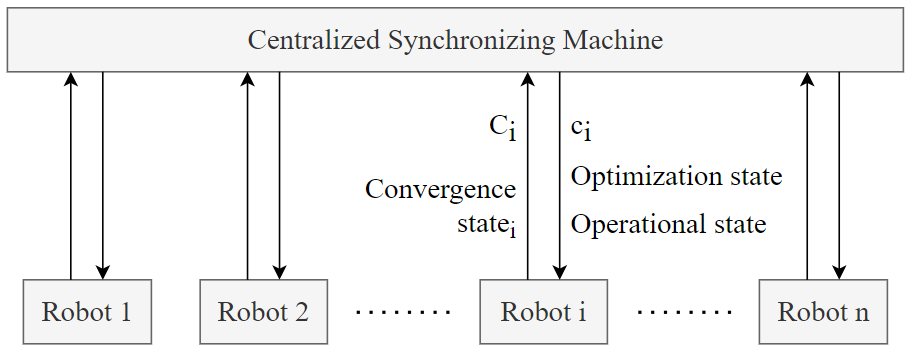}
    \setlength{\belowcaptionskip}{-15pt}  
    \caption{System Architecture}
    \label{fig:System Architecture}
\end{figure}

Fig.~\ref{fig:System Architecture} shows the architecture of the coverage control system. There are two types of component in the system, including a centralized synchronizing machine and a team of $n$ robots. The synchronizing machine is used to synchronize all the robots and the robots are used to cover the environment and perform patrolling or monitoring tasks. It is assumed that limited communication between the synchronizing machine and robots is already established.

\subsubsection{Centralized Synchronizing Machine}
The purpose of the centralized synchronizing machine is to synchronize all the robots, including their normalized capabilities $c$, the optimization state, and the operational state. \par
The normalized capabilities $c = \{ c_1, c_2, \ldots, c_n \} \in \mathbb{R}^n_{\geq 0}$ will be used in the optimization. It is obtained by $l_1$-normalizing the capabilities of all robots. Since the capabilities are non-negative, the $l_1$-normalization of the capabilities are as following.
\begin{equation}
    \label{eqn:norm. cap.}
    c_i = \frac{C_i}{\sum_{k=1}^n C_k} \quad \forall i \in \mathcal{N}
\end{equation} \par
The computation of the normalized capability $c_i$ takes the capabilities of all robots, which might vary over time due to many reasons like decreasing in robots' remaining battery power or gain/removal of robots. To maintain the accuracy of $c$, the synchronizing machine keeps receiving the capabilities of all robots $C = \{C_1,C_2,\ldots,C_n\}$, and sends the corresponding $c_i$ to each robot $i$. \par
The partition of $Q$ depends on the generators' positions $\mathcal{P}$ and weights $\mathbf{w}$. The Block Coordinate Descent (BCD) method is used to optimize two blocks of variables alternately. Hence, a synchronized optimization state on which blocks to optimize is needed for all robots. The optimization state will toggle if the synchronizing machine received convergence signals from all robots or the run time is too long. \par
The operational state is the instruction sent to all robots in order to synchronize the robots' operation. The operational state can be ``initialing" or ``patrolling". The system will start from initialing state, and it will change to patrolling state if all robots' generators converge in both positions and weights. \par

\subsubsection{Robots}
The task of the robots is to patrol or monitor a known environment. All robots are assumed to be equipped with a good positioning system and actuation system so that they know their global positions and they are capable to move to any target position $\hat{x}_i \in Q$. Besides the communication with the synchronizing machine, robots are assumed to be able to communicate locally with their neighbours to exchange the information of $A_i$, $c_i$, $p_i$, and $w_i$. With all the neighbour generators' positions and weights, robot $i$ is capable to find $W_i$ and calculate its cell area $A_i$ by the Shoelace formula \cite{Shoelace formula}. \par
Depending on the optimization state from the synchronizing machine, robots will compute the gradient and update their generator's positions or weights in a distributed manner. It will send a convergence signal to the synchronizing machine if the change in the optimization variable is small. \par
Behaviors of robots are depending on the operational state received from the synchronizing machine. The target position of any robot $i$ is designed as follow,
\begin{equation}
    \label{eqn:target position}
    \hat{x}_i = \
    \begin{cases}
        p_i & \text{if operational state =``initialing"}   \\
        patrol(W_i) & \text{if operational state = ``patrolling"}
    \end{cases} 
\end{equation} \par
When the operational state is ``initialing", $\hat{x}_i$ follows $p_i$ to move to its optimal cell position. When the operational state is ``patrolling", $\hat{x}_i$ updates to patrol its allocated cell $W_i$ by following any single cell coverage path planning algorithms, for example CPP algorithms in Cabreira et al. study~\cite{Cabreira}. \par


\section{Coverage Control Method} \label{section:method}

\subsection{Space Partitioning Method} 
In this sub-section, two important partitioning methods, Voronoi diagram partition and power diagram partition, will be discussed. Cells of robots $\mathcal{W}$ are generated by their generators' positions $\mathcal{P}$ and weights $\mathbf{w}$ under these methods. Generator $i$ and generator $j$ are said to be neighbour if and only if their corresponding cells share a common edge, i.e., $\partial W_i \cap \partial W_j \neq \emptyset$ and $i \neq j$. The collection of all neighbours of generator $i$ is defined as $\mathcal{N}_i \subset \mathcal{N}$.

\subsubsection{Voronoi Diagram Partition} 
The cell corresponding to generator $i$ under Voronoi diagram partitioning method is defined as $W_i = \{ q \in Q \ | \ d(p_i,q) \leq d(p_j,q) , \ \forall j \in \mathcal{N}_i \}$, where $d(a,b) = \norm{a-b}$. It can be computed by the positions of its generator and neighbour generators. All cells in Voronoi diagram are non-empty and the generators are always located in their corresponding cell. As shown in fig.~\ref{fig:Voronoi}, the common edge of two adjacent cells $W_i$ and $W_j$ is a segment of the perpendicular bisector of $s_{ij}$, where $s_{ij}$ is the segment from $p_i$ to $p_j$. It can be formulated as
\begin{align}    
    \label{eqn:Voronoi_edge}
        & \partial W_i \cap \partial W_j = \{ q \in Q \ | \ q = m + dt , \ t \in [t_{min}, t_{max}] \} \nonumber \\
        & \text{with} \quad m = \frac{1}{2} (p_i + p_j), \ d = R_{\pi/2}\frac{p_j-p_i}{\norm{p_j-p_i}} , 
\end{align}
where $R_{\pi/2}$ is the rotation matrix of $\frac{\pi}{2}$, and $t_{min}$, $t_{max}$ can be computed from the vertices of its cell. 

\subsubsection{Power Diagram Partition}
\begin{figure}[t]
    \centering
        \begin{subfigure}[h]{0.235\textwidth}
            \includegraphics[width=\textwidth]{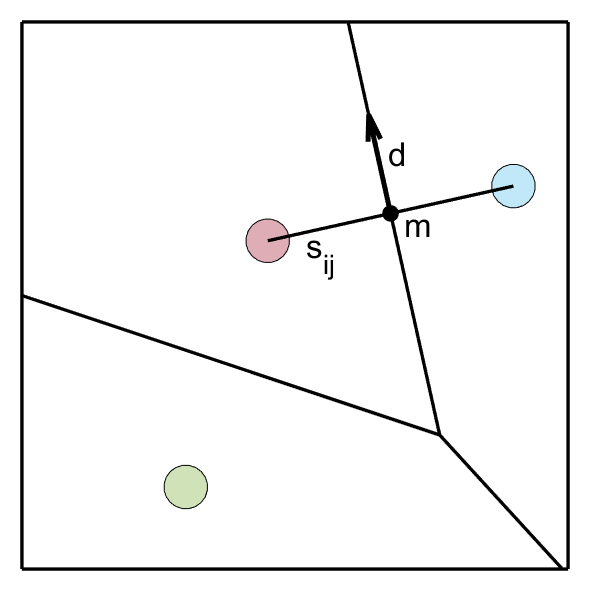}
            \caption{Voronoi Diagram}
            \label{fig:Voronoi}
        \end{subfigure}
            \begin{subfigure}[h]{0.235\textwidth}
            \includegraphics[width=\textwidth]{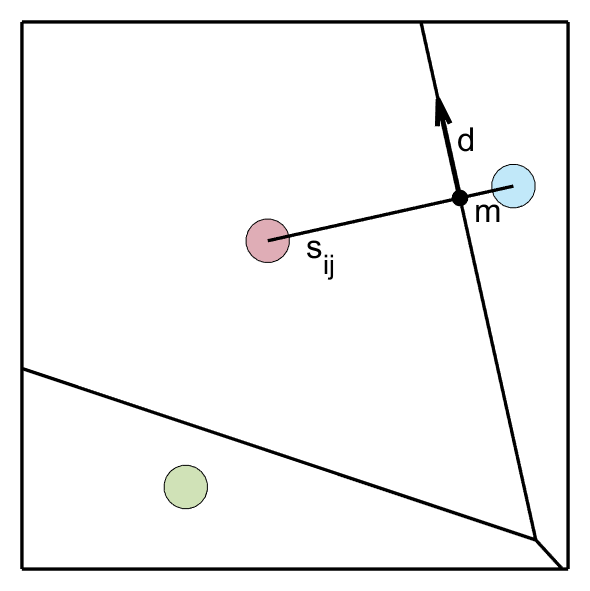}
            \caption{Power Diagram}
            \label{fig:Power}
        \end{subfigure}
    \setlength{\belowcaptionskip}{-15pt}  
    \caption{An example of Voronoi Diagram and Power Diagram}
    \label{fig:Voronoi and Power}
\end{figure}

The power diagram partitioning method is the generalized version of the Voronoi diagram partitioning method by introducing the additional weights $\mathbf{w}$ to the generators. The cell corresponding to generator $i$ is defined as $W_i = \{ q \in Q \ | \ d(p_i,q,w_i) \leq d(p_j,q,w_j) , \ \forall j \in \mathcal{N}_i \}$, where $d(a,b,w) = \norm{a-b} - w$. It can be computed by the positions and weights of its generator and neighbour generators. Unlike Voronoi diagram, cells in power diagram can be empty and generators might be located outside of their corresponding cell. An example is shown in Fig.~\ref{fig:Power}. The common edge of two adjacent cells $W_i$ and $W_j$ might not be coincident with the mid-point of $s_{ij}$, but it is still perpendicular to $s_{ij}$. The edge is formulated as
\begin{align} 
    \label{eqn:Power_edge}
        \partial W_i \cap \partial W_j & = \{ q \in Q \ | \ q = m + dt , \ t \in [t_{min}, t_{max}] \} \nonumber \\
        \text{with} \quad  m & = \frac{1}{2} (p_i + p_j) + \frac{w_i-w_j}{2 \| p_j-p_j \|_2} (p_j - p_i) \nonumber \\
        d & = R_{\pi/2}\frac{p_j-p_i}{\|p_j-p_i\|_2} , 
\end{align}
where $R_{\pi/2}$ is the rotation matrix of $\frac{\pi}{2}$, and $t_{min}$, $t_{max}$ can be computed from the vertices of its cell. The power diagram is equivalent to the Voronoi diagram if and only if the weights of all generators are equal, i.e., $w_i = w_j \ , \ \forall i,j \in \mathcal{N}$.

\subsection{Optimization}
To balance the area allocated to each robot, the target ratios of $A_i$ over the total area $A$ is $c_i$ since $\sum_{i=1}^n c_i = 1$, which gives $c_i = \frac{A_i}{A} \ , \ \forall i \in \mathcal{N}$. After some manipulations of terms, the error of robot $i$ is scaled to the error with the total area $A$ and defined as $\frac{A_i}{c_i} - A \ , \ \forall i\in \mathcal{N}$. The objective function $H$ for the whole system to minimize is designed as the sum of square errors of all robots as follows.
\begin{equation}
    \label{eqn:objective function}
    H = \sum_{i=1}^n (\frac{A_i}{c_i} - A)^2
\end{equation} \par
Given $c_i \ \forall i \in \mathcal{N}$ and $A$, $H$ is a function of the cell area $A_i \ \forall i \in \mathcal{N}$. The partition of cells $\mathcal{W}$ use the power diagram partitioning method shown in section III-A2, which can be found by $\mathcal{P}$ and $\mathbf{w}$. Hence, $H$ is a function of $\mathcal{P}$ and $\mathbf{w}$, which can be written as $H(\mathcal{P}, \mathbf{w})$. \par
At the beginning, $p_i$ is set as the robot initial position $x_i \in Q$ and $w_i$ is set to zero $\forall i \in \mathcal{N}$. Therefore $\mathcal{W}$ follows the Voronoi diagram partitioning method described in section III-A1 and all the generators are located in their corresponding cells. Since there are two blocks of variables, $\mathcal{P}$ and $\mathbf{w}$, can be chosen as the optimization variable for minimizing $H$, the block coordinate descent (BCD) method is used to optimize $H$ alternately as follow,
\begin{align}
    \label{eqn:BCD}
        \text{Position optimization: } & \mathcal{P}^{k+1} = arg \min_{\mathcal{P} \in \mathcal{W}} \ H(\mathcal{P}^k, \mathbf{w}^k) \nonumber \\
        \text{Partition optimization: } & \mathbf{w}^{k+1} = arg \min_{\mathbf{w}} \ H(\mathcal{P}^{k+1}, \mathbf{w}^k) \nonumber \\
        & \text{s.t. } \mathcal{P} \in \mathcal{W},
\end{align}
where $k$ is the number of iterations of the BCD method. \par
After all generators' positions $\mathcal{P}$ and weights $\mathbf{w}$ converged, robots start to patrol their allocated cell while $p_i$ and $w_i$ keep optimizing $\forall i \in \mathcal{N}$ to deal with the dynamic changing of robots' capabilities. Since $\mathcal{P}$ is always lies within $\mathcal{W}$ and all robots follow their corresponding generator's position before patrolling, robots should start their patrolling task in their allocated cells when they switch to patrolling state. Therefore, the patrolling task can be considered as multiple single robot patrolling tasks and those do not need to consider the behavior of the others. \par
Gradient descent based method is used for both optimizations in $\mathcal{P}$ and $\mathbf{w}$. The optimization variables will be updated according to the gradients and the gradients can be computed in a distributed manner. \par

\subsubsection{Position Optimization}
The position optimization optimize the position of the generator $p_i$ for each robot $i$. The change on generators' positions follows
\begin{equation}
    \label{eqn:pos update}
    \Dot{p}_i = u_{p,i} \quad \forall i \in \mathcal{N},
\end{equation}
where $u_{p,i}$ is the gradient based descent update for the position of generator $i$. \par
Consider the gradient of the objective function $H$ with respect to the generator position $p_i$. Since any small step change in $p_i$ can only affect $A_i$ and $A_j \ \forall j \in \mathcal{N}_i$, the gradient $\frac{\partial H}{\partial p_i}$ can be simplified as,
\begin{equation}
    \label{eqn:dh dp}
    \begin{split}
        \frac{\partial H}{\partial p_i} 
        & = 2(\frac{A_i}{c_i} - A)(\frac{1}{c_i}) \frac{\partial A_i}{\partial p_i} + \sum_{j \in \mathcal{N}_i} 2(\frac{A_j}{c_j} - A)(\frac{1}{c_j}) \frac{\partial A_j}{\partial p_i} \\
    \end{split}
\end{equation} \par
The derivatives of the areas $A_i$ and $A_j$ can be written as,
\begin{equation}
    \label{eqn:dAi dpi}
    \frac{\partial A_i}{\partial p_i} = \int_{\partial W_i} n_i^T(q) \frac{\partial q}{\partial p_i} dq = \sum_{l \in \mathcal{N}_i} \int_{\partial W_i \cap \partial W_l} n_i^T(q) \frac{\partial q}{\partial p_i} dq \\
\end{equation}
\begin{equation}
    \label{eqn:dAj dpi}
    \frac{\partial A_j}{\partial p_i} = \int_{\partial W_j} n_j^T(q) \frac{\partial q}{\partial p_i} dq = \sum_{l \in \mathcal{N}_j} \int_{\partial W_j \cap \partial W_l} n_j^T(q) \frac{\partial q}{\partial p_i} dq , \\
\end{equation}
where $n_i(q) = \frac{p_j - p_i}{\norm{p_j-p_i}} $ is the outward normal vector of $W_i$, evaluated at any point $q \in \partial W_i \cap \partial W_j, \ j \in \mathcal{N}_i$. \par
For cell $W_j$ with $j \in \mathcal{N}_i$, small change in $p_i$ will only affect the common edge with $W_i$, i.e. $\frac{\partial q}{\partial p_i} = 0 \ \forall q \in \ \partial W_j \cap \partial W_l \text{ and } l \neq i$. In addition, $n_i(q)$ equal to $-n_j(q)$ for any point $q \in \partial W_i \cap \partial W_j$. As a result, \eqref{eqn:dAj dpi} can be simplified and
the gradient of $H$ is expressed as
\begin{align}
    \label{eqn:dh dp 2}   
        & \frac{\partial H}{\partial p_i} = 2(\frac{A_i}{c_i} - A)(\frac{1}{c_i}) \sum_{l \in \mathcal{N}_i}D_{il} - \sum_{j \in \mathcal{N}_i} 2(\frac{A_j}{c_j} - A)(\frac{1}{c_j}) D_{ij} , \nonumber \\
        & \text{where} \quad D_{ij} = \int_{\partial W_i \cap \partial W_j} n_i^T(q) \frac{\partial q}{\partial p_i} dq \quad \forall i \in \mathcal{N}
\end{align}
Here $D_{ij}$ is the change in $A_i$ with respect to $p_i$ by considering the common edge $\partial W_i \cap \partial W_j$. \par
From \eqref{eqn:Power_edge}, points on the power diagram cells' edge can be expressed in terms of $t \in [t_{min}, t_{max}]$. By differentiating \eqref{eqn:Power_edge} with respect to $p_i$, we obtain
\begin{align*}
        \frac{\partial q}{\partial p_i} & = \frac{1}{2}I + \frac{w_i-w_j}{2\norm{p_j - p_i}^4} \Psi + \frac{t}{\norm{p_j - p_i}^3} \Omega , \nonumber \\
\end{align*}
where
\begin{align}
        & \Psi = 
        \begin{bmatrix}
            (x_j-x_i)^2 - (y_j-y_i)^2  &  2(x_j-x_i)(y_j-y_i) \\
            2(x_j-x_i)(y_j-y_i)  &  (y_j-y_i)^2 - (x_j-x_i)^2 \\
        \end{bmatrix}, \nonumber \\
        & \Omega = 
        \begin{bmatrix}
            -(x_j-x_i)(y_j-y_i)  &  (x_j-x_i)^2 \\
            -(y_j-y_i)^2  &  (x_j-x_i)(y_j-y_i) \\
        \end{bmatrix}, \nonumber \\
        & p_i = 
        \begin{bmatrix}
            x_i \\ y_i \\
        \end{bmatrix} 
        , \quad
        p_j = 
        \begin{bmatrix}
            x_j \\ y_j \\
        \end{bmatrix} 
        , \quad \forall q \in \partial W_i \cap \partial W_j 
\end{align} \par
Finally, $D_{ij}$ is rewritten as
\begin{align}
    \label{eqn:Dij}
        & D_{ij} = \int_{t_{min}}^{t_{max}} n_i^T(q) \frac{\partial q}{\partial p_i} dt \nonumber \\
        & = ( \frac{1}{2\norm{p_j - p_i}} + \frac{w_i-w_j}{2\norm{p_j - p_i}^3}) (t_{max}-t_{min})
        \begin{bmatrix}
            x_j-x_i \\ y_j-y_i
        \end{bmatrix}^T \nonumber \\
        & \quad  + \frac{t_{max}^2-t_{min}^2}{\norm{p_j - p_i}^2}
        \begin{bmatrix}
            -(y_j-y_i) \\ x_j-x_i
        \end{bmatrix}^T 
        \quad \forall i,j \in \mathcal{N}
\end{align} 
\quad Equation \eqref{eqn:dh dp 2} together with \eqref{eqn:Dij} show the computation of gradient with respect to $p_i$, which can be computed in a distributed manner. To obtain the descent update $u_{p,i}$ for generator $i$, the gradient of $H$ with respect to $p_i$ is multiplied by $-\gamma_{p,i}$, where $\gamma_{p,i} \in \mathbb{R}_{\geq 0} \ \forall i \in \mathcal{N}$. It maintains a small step for the descent and prevents the gradient from being too sensitive by stopping the generators moving too close with each other. The resulting vector will be projected to the boundary of the corresponding cell if it is moving outward to maintain $\mathcal{P} \in \mathcal{W}$. Finally, the position descent update is
\begin{align}
        & u_{p,i} = 
        \begin{cases}
            proj(\tilde{u}_{p,i}) & \text{if $p_i \in \partial W_i$ $\land$ $\tilde{u}_{p,i}$ pointing out} \\
            \tilde{u}_{p,i} & \text{otherwise} \\
        \end{cases} \nonumber \\
        & \forall i \in \mathcal{N}, \nonumber \\ 
        & \text{where} \quad \tilde{u}_{p,i} = -\gamma_{p,i}\frac{\partial H}{\partial p_i}^T 
\end{align}

\subsubsection{Partition Optimization}
The partition optimization optimize the weight of the generator $w_i$ for each robot $i$. The update on generators' weights are as following.
\begin{equation}
    \label{eqn:weight update}
    \Dot{w_i} = u_{w,i} \quad \forall i \in \mathcal{N},
\end{equation}
where $u_{w,i}$ is the descent update for weight of generator $i$. \par
By similar arguments in section III-B1, the gradient of $H$ with respect to the generator weight $w_i$ has the form
\begin{equation}
    \label{eqn:dh dw}
        \frac{\partial H}{\partial w_i} = 2(\frac{A_i}{c_i} - A)(\frac{1}{c_i}) \frac{\partial A_i}{\partial w_i} + \sum_{j \in \mathcal{N}_i} 2(\frac{A_j}{c_j} - A)(\frac{1}{c_j}) \frac{\partial A_j}{\partial w_i} \\
\end{equation}
and the derivatives of the $A_i$ and $A_j$ with respect to $w_i$ is
\begin{equation}
    \label{eqn:dAi dwi}
    \frac{\partial A_i}{\partial w_i} = \sum_{l \in \mathcal{N}_i} \int_{\partial W_i \cap \partial W_l} n_i^T(q) \frac{\partial q}{\partial w_i} dq \\
\end{equation}
\begin{equation}
    \label{eqn:dAj dwi}
    \frac{\partial A_j}{\partial w_i} = - \int_{\partial W_i \cap \partial W_j} n_i^T(q) \frac{\partial q}{\partial w_i} dq 
\end{equation} \par
Hence, the gradient of $H$ is simplified to
\begin{align}
    \label{eqn:dh dw 2}   
        & \frac{\partial H}{\partial w_i} = 2(\frac{A_i}{c_i} - A)(\frac{1}{c_i}) \sum_{l \in \mathcal{N}_i}E_{il} - \sum_{j \in \mathcal{N}_i} 2(\frac{A_j}{c_j} - A)(\frac{1}{c_j}) E_{ij} , \nonumber \\
        & \text{where} \quad E_{ij} = \int_{\partial W_i \cap \partial W_j} n_i^T(q) \frac{\partial q}{\partial w_i} dq \quad \forall i \in \mathcal{N}
\end{align}
Here $E_{ij}$ is the change in $A_i$ with respect to $w_i$ by considering the common edge $\partial W_i \cap \partial W_j$. \par
By differentiating \eqref{eqn:Power_edge} with respect to $w_i$, we obtain
\begin{equation}
    \frac{\partial q}{\partial w_i} = \frac{1}{2\norm{p_j - p_i}}n_i(q) \quad \forall q \in \partial W_i \cap \partial W_j
\end{equation} \par
Hence, $E_{ij}$ is rewrote as
\begin{equation}
    \label{eqn:Eij}
    \begin{split}
        E_{ij} & = \int_{t_{min}}^{t_{max}} n_i^T(q) \frac{\partial q}{\partial w_i} dt 
        = \frac{t_{max}-t_{min}}{2\norm{p_j - p_i}} \quad \forall i,j \in \mathcal{N}
    \end{split}
\end{equation} \par
Equation \eqref{eqn:dh dw 2} together with \eqref{eqn:Eij} show the computation of the gradient with respect to $w_i$, which can also be computed in a distributed manner. It is then multiplied by $-\gamma_{w,i}$ to produce the descent update $u_{w,i}$ on the weight of generator $i$, where $\gamma_{w,i} \in \mathbb{R}_{\geq 0} \ \forall i \in \mathcal{N}$. $-\gamma_{w,i}$ is used for maintaining a small step and keep $\mathcal{P} \in \mathcal{W}$. As a result, \par
\begin{equation}
    u_{w,i} = -\gamma_{w,i} \frac{\partial H}{\partial w_i} \quad \forall i \in \mathcal{N}
\end{equation}

\subsubsection{Convergence}
Consider the derivative of the objective function $H$ with respect to time $t$, 
\begin{equation}
    \label{eqn:dH dt}
    \frac{dH(\mathcal{P},\mathbf{w})}{dt} = \sum_{i=1}^n \frac{\partial H}{\partial p_i} \frac{dp_i}{dt} + \sum_{i=1}^n \frac{\partial H}{\partial w_i} \frac{dw_i}{dt}
\end{equation} \par
At position optimization, $\frac{dp_i}{dt} = u_{p,i}$ and $\frac{dw_i}{dt} = 0$ $\forall i \in \mathcal{N}$. The derivative of $H$ is 
\begin{align*}
    \frac{dH(\mathcal{P},\mathbf{w})}{dt} 
    & = \sum_{i=1}^n \frac{\partial H}{\partial p_i} [\sigma_i proj(\tilde{u}_{p,i}) + (1-\sigma_i)(\tilde{u}_{p,i})] \\
    & = \sum_{i=1}^n -\gamma_{p,i} \norm{\frac{\partial H}{\partial p_i}}^2 [\sigma_i \cos{\theta_i}+(1-\sigma_i)] ,
\end{align*}
where
\begin{align}
    & \sigma_i = 
    \begin{cases}
        1 & \quad \text{if projection needed}   \\
        0 & \quad \text{otherwise}
    \end{cases} \nonumber
    \quad \forall i \in \mathcal{N} \\
    & \theta_i \in [0,\frac{\pi}{2}] \text{ is the angle between boundary and $\tilde{u}_{p,i}$}
\end{align}
At partition optimization, $\frac{dp_i}{dt} = 0$ and $\frac{dw_i}{dt} = u_{w,i}$ $\forall i \in \mathcal{N}$. The derivative of $H$ is
\begin{equation}
    \label{eqn:dH dt for part.}
    \frac{dH(\mathcal{P},\mathbf{w})}{dt} = \sum_{i=1}^n \frac{\partial H}{\partial w_i} (-\gamma_{w,i} \frac{\partial H}{\partial w_i}) 
    = \sum_{i=1}^n -\gamma_{w,i} \norm{\frac{\partial H}{\partial w_i}}^2
\end{equation} \par
Combining both position and partition optimization, the derivative of $H$ for the whole system can be expressed as,
\begin{align}
    \label{eqn:dH dt for whole}
            & \frac{dH(\mathcal{P},\mathbf{w})}{dt} = \lambda \sum_{i=1}^n -\gamma_{p,i} \norm{\frac{\partial H}{\partial p_i}}^2 [\sigma_i \cos{\theta_i}+(1-\sigma_i)] \nonumber \\
            & \qquad \qquad \qquad + (1-\lambda)\sum_{i=1}^n -\gamma_{w,i} \norm{\frac{\partial H}{\partial w_i}}^2 , \nonumber \\
        & \text{where} 
        \quad \lambda = 
        \begin{cases}
            1 & \quad \text{at position optimization}   \\
            0 & \quad \text{at partition optimization}
        \end{cases}
\end{align} \par
Since $\gamma_{p,i}, \gamma_{w,i} \in \mathcal{R}_{\geq 0}$ and $\cos{\theta_i}\in[0,1] \ \forall i \in \mathcal{N}$, the derivative of $H$ with respect to time is always smaller or equal to zero $\forall \ \mathcal{P},\mathbf{w}$. The objective function $H$ decreases monotonically to a local minimum. \par

We summarize the above procedures in the following two algorithms, where $Transmit()$, $Receive()$ control the communication between the synchronizing machine and robot, $receive()$ control the communication between robots locally, $PositionOptimization()$ and $PartitionOptimization()$ find the descent update by the calculation shown above, and $Patrol()$ returns the target position to patrol the cells.
\begin{algorithm}                     
\caption{An algorithm for each robot $i$}\label{alg:robot}        
\begin{algorithmic}[1]                   
\Loop
    \State $C_i \gets UpdateCapability()$
    \State $Transmit(C_i)$
    \State $c_i \gets Receive(NormalizedCapability)$
    \State $\lambda \gets Receive(OptimizationState)$
    \ForEach {$j \in \mathcal{N}_i$}
        \State $c_j \gets receive(NormalizedCapability)$
        \State $A_j \gets receive(Area)$         
    \EndFor
    \State $W_i = ComputeCell()$
    \State $A_i = ComputeCellArea()$
    \If{$\lambda = 1$}
        \State $u_{p,i} \gets PositionOptimization()$
        \State $p_i \gets p_i + u_{p,i}$
        \If{$ \norm{u_{p,i}} < PositionUpdateTheshold $}
            \State $Transmit(converged)$
        \EndIf
    \ElsIf{$\lambda = 0$}
        \State $u_{w,i} \gets PartitionOptimization()$
        \State $w_i \gets w_i + w_{w,i}$
        \If{$|u_{w,i}| < WeightUpdateTheshold $}
            \State $Transmit(converged)$
        \EndIf
    \EndIf         
    \\
    \State $OperationState \gets Receive(OperationState)$
    \If{$OperationState = ``initialing"$}
        \State $\hat{x}_i \gets p_i$
    \ElsIf{$OperationState = ``Patrolling"$}
        \State $\hat{x}_i \gets Patrol()$
    \EndIf     
\EndLoop
\end{algorithmic}
\end{algorithm}


\begin{algorithm}
\caption{An algorithm for the synchronizing machine}\label{alg:sync mach}
\begin{algorithmic} [1]
\State $\lambda \gets 1$        \Comment{Start with position optimization}
\State $time \gets 0$
\State $keepConverged \gets 0$
\Loop
    \ForEach {$i \in \mathcal{N}$}
        \State $C_i \gets Receive(Capability)$
    \EndFor
    \State $c = ComputeNormalizedCapabilities()$
    \ForEach {$i \in \mathcal{N}$}
        \State $Transmit(c_i, \ \lambda)$
    \EndFor
    \\
    \If{Received converged from all robots}
        \State $\lambda \gets \mod{(\lambda+1, 2)}$
        \State $keepConverged \gets keepConverged +1 $
        \State $time \gets 0$
    \ElsIf{$time \geq timeThreshold$}
        \State $\lambda \gets \mod{(\lambda+1, 2)}$
        \State $keepConverged \gets 0$
        \State $time \gets 0$
    \Else
        \State $time \gets time + 1$
    \EndIf
    
    \If{$keepConverged \geq 2$}
        \State $Transmit(``Patrolling")$
    \Else
        \State $Transmit(``Initialing")$
    \EndIf
\EndLoop
    
\end{algorithmic}
\end{algorithm}

\newcommand\imgSizeScale{0.28}
\begin{figure*}[t]
    \centering
    \begin{subfigure}[t]{\imgSizeScale \textwidth}
        \includegraphics[width=\textwidth]{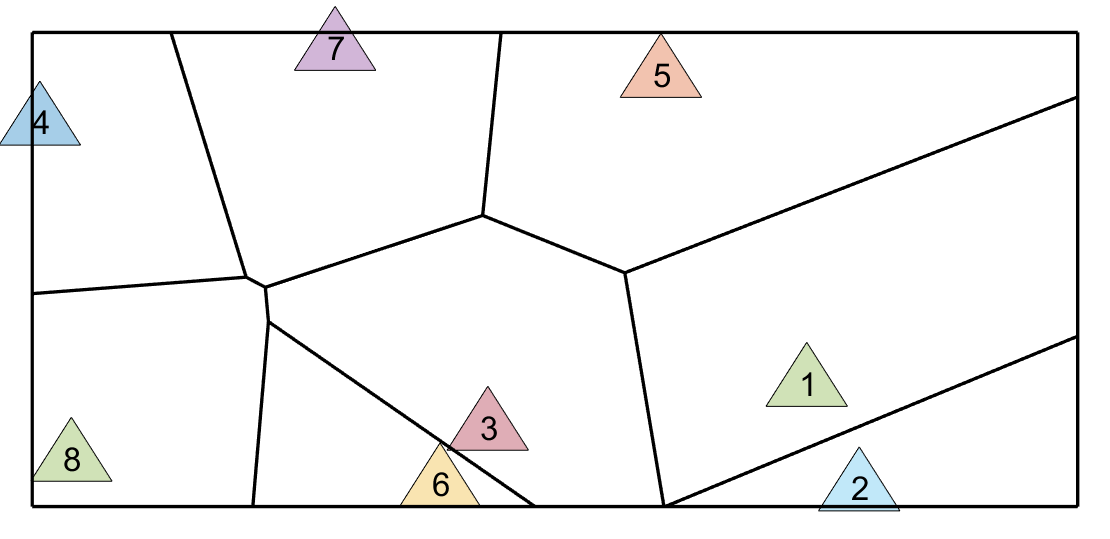}
        \caption{Initial setting}
        \label{fig:init_setting}
    \end{subfigure}
    \begin{subfigure}[t]{\imgSizeScale \textwidth}
        \includegraphics[width=\textwidth]{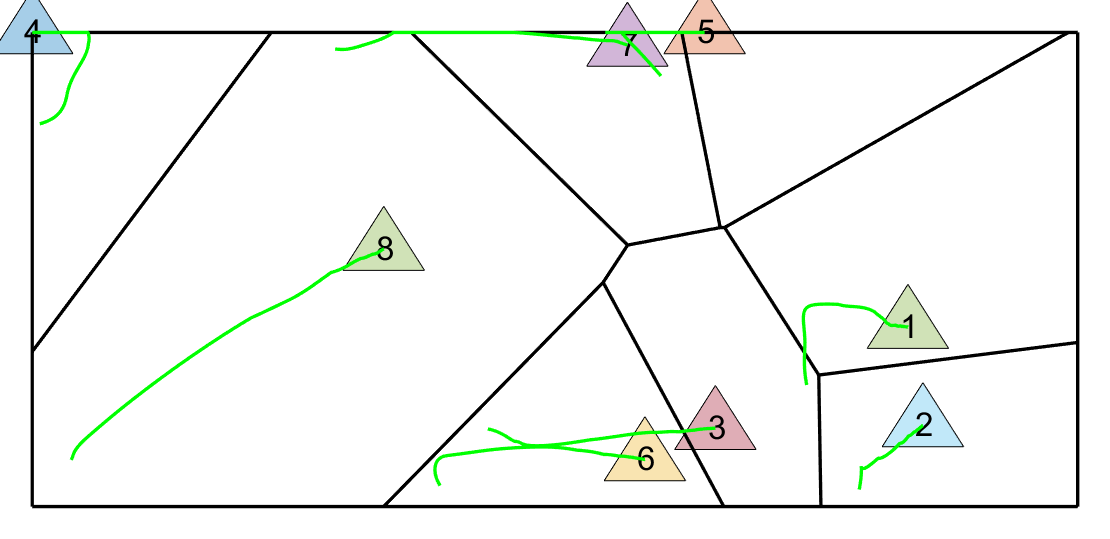}
        \caption{The first convergence}
        \label{fig:first convergence}
    \end{subfigure}
    \begin{subfigure}[t]{\imgSizeScale \textwidth}
        \includegraphics[width=\textwidth]{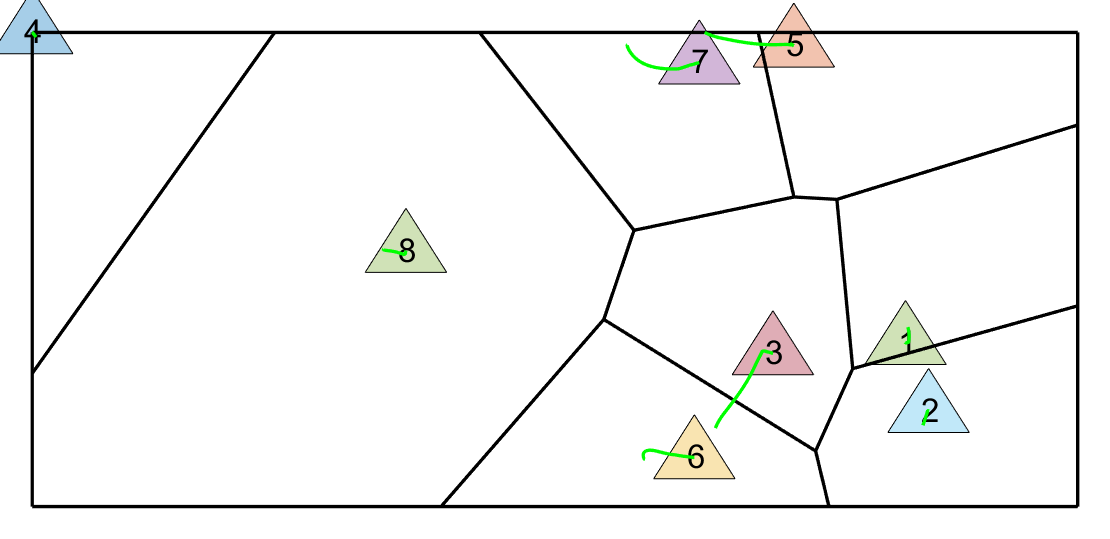}
        \caption{Second convergence}
        \label{fig:second convergence}
    \end{subfigure}
    \par
    \begin{subfigure}[t]{\imgSizeScale \textwidth}
        \includegraphics[width=\textwidth]{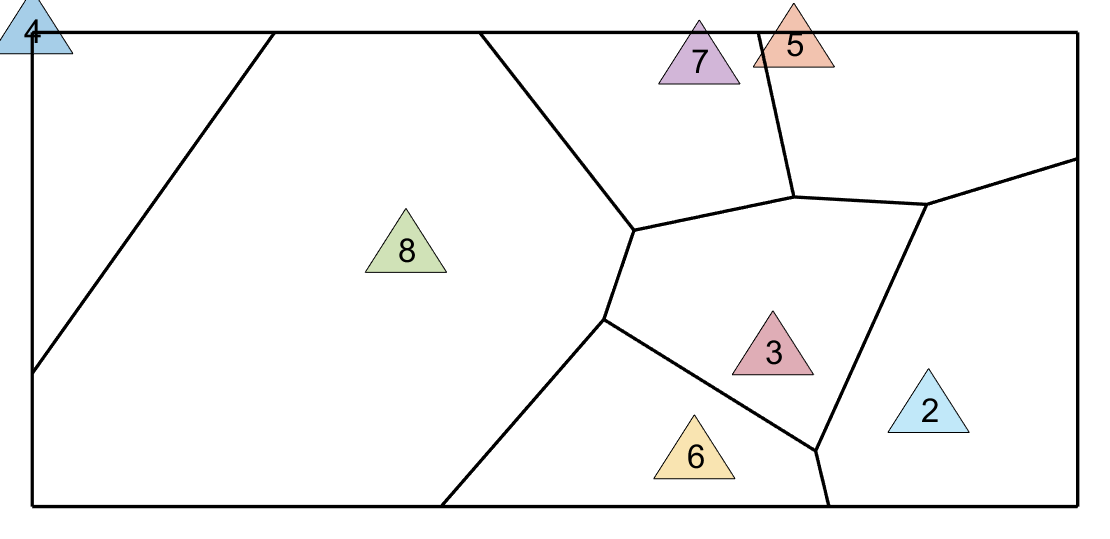}
        \caption{After removing robot 1}
        \label{fig:removed R1}
    \end{subfigure}
    \begin{subfigure}[t]{\imgSizeScale \textwidth}
        \includegraphics[width=\textwidth]{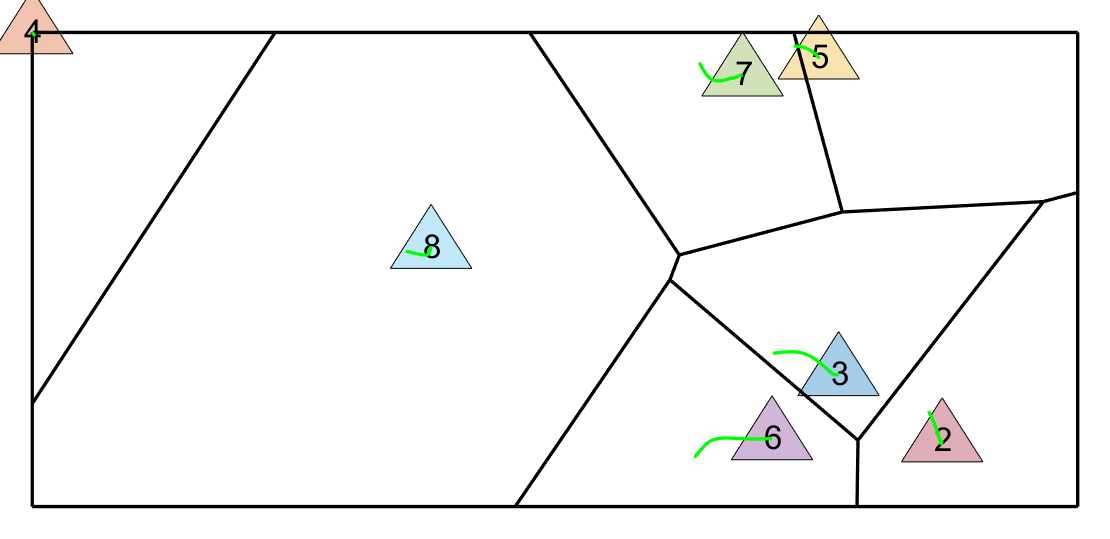}
        \caption{Third convergence}
        \label{fig:third convergence}
    \end{subfigure}
    \begin{subfigure}[t]{\imgSizeScale \textwidth}
        \includegraphics[width=\textwidth]{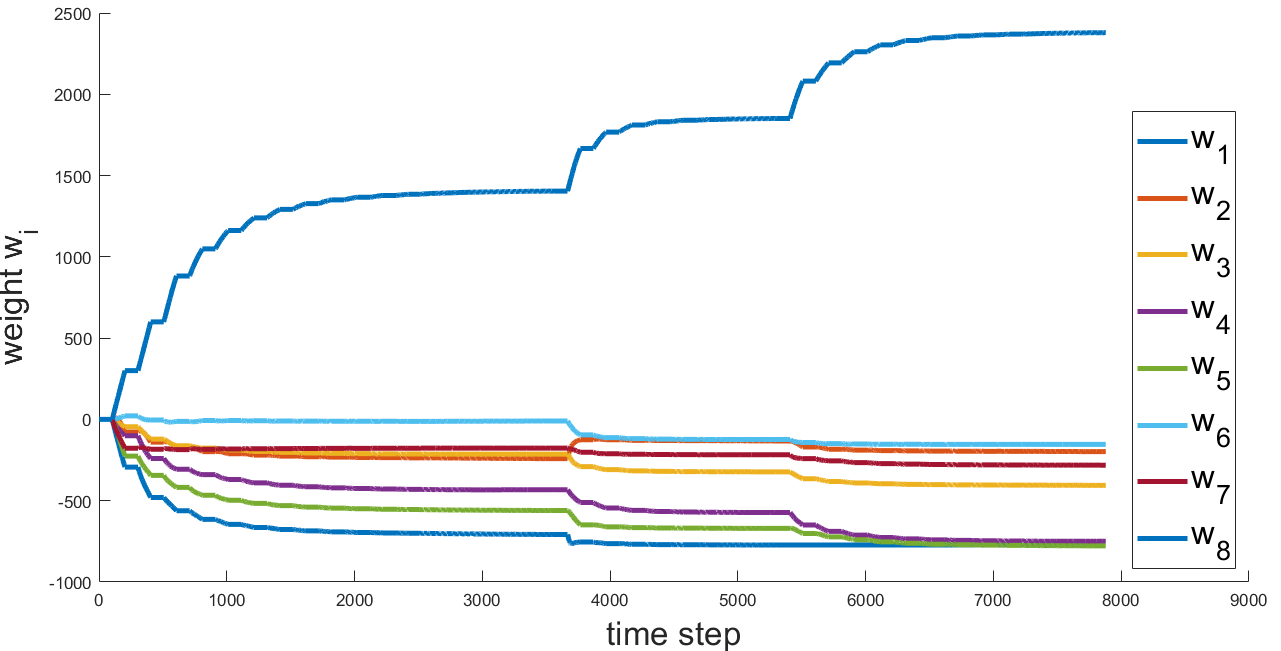}
        \caption{The weight $\mathcal{W}$}
        \label{fig:weight}
    \end{subfigure}
    \setlength{\belowcaptionskip}{-15pt}  
    \caption{Simulation results on generators' position and weight}
    \label{fig:Simulation result 1}
\end{figure*}

\begin{figure}[t]
    \centering
    \setlength{\belowcaptionskip}{-5pt}   
    \includegraphics[width=\imgSizeScale \textwidth]{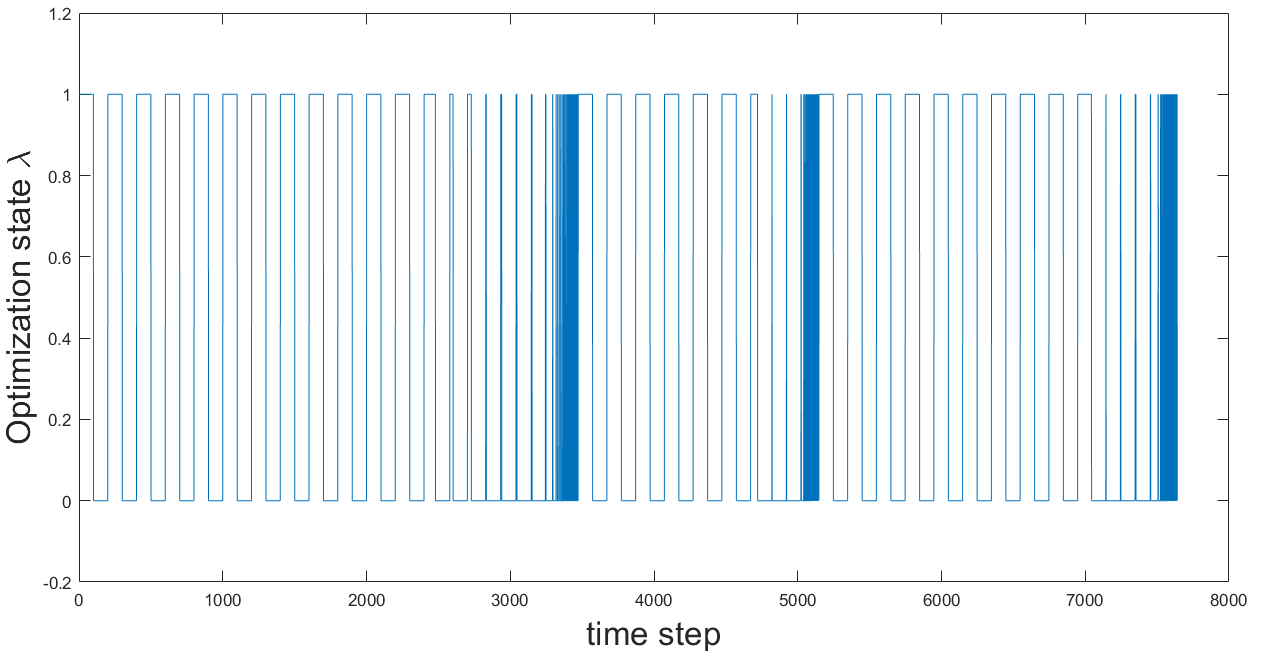}
    \caption{The optimization state $\lambda$}
    \label{fig:opt_state}
\end{figure}
\begin{figure}[t]
    \centering
    \setlength{\belowcaptionskip}{-15pt}   
    \includegraphics[width=\imgSizeScale \textwidth]{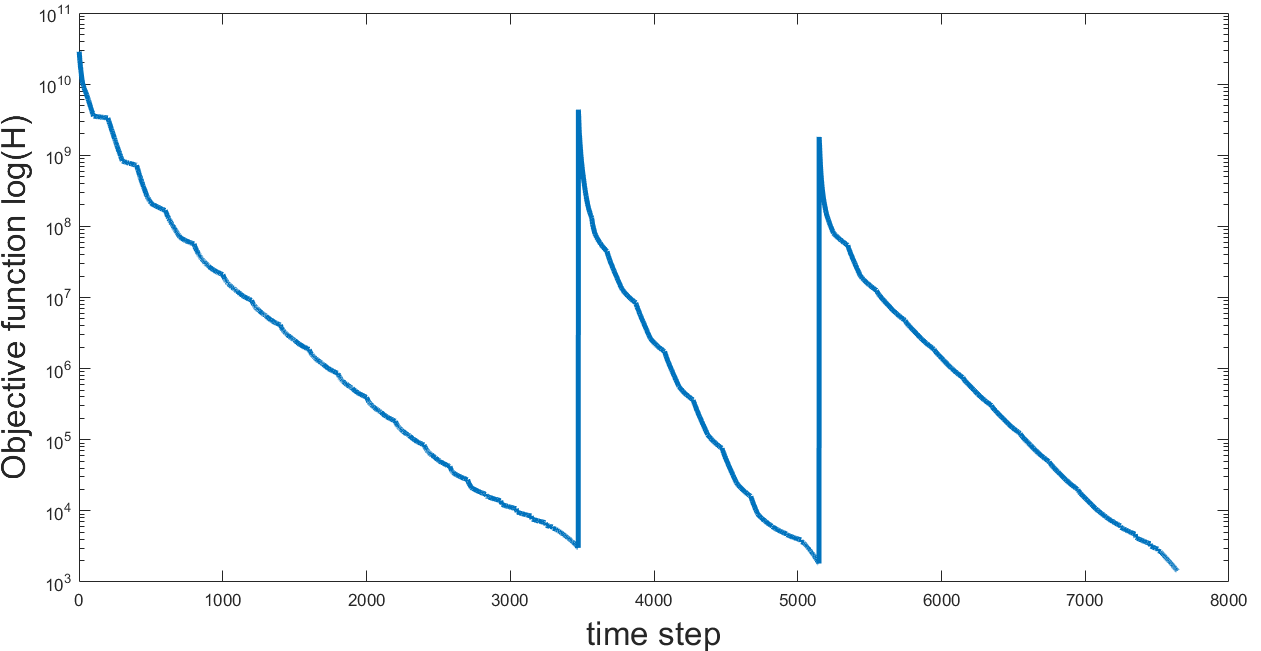}
    \caption{The objective function $H$ value in log scale}
    \label{fig:obj_val log}
\end{figure}
\section{Simulation} \label{section:simulation}
In this section, Matlab simulation results of the proposed coverage control system are provided and discussed. \par
Fig.~\ref{fig:init_setting} shows the initial setting of generators' positions which are randomly placed in a rectangular environment with vertices $(0,0), (200,0), (200,100)$ and $(0,100)$. The capabilities of robots are [2 1 1 1 1 1 1 5]. The system starts to run and converges at the configuration in Fig.~\ref{fig:first convergence}. Then the capability of robot 1 is halved to simulate the dynamic changes of robots' capabilities. The system continues and re-converges to the configuration in Fig.~\ref{fig:second convergence}. Fig.~\ref{fig:removed R1} simulates the situation that robot 1 is lost due to some reasons and Fig.~\ref{fig:third convergence} shows the re-converged configuration with seven robots. The evolution of weights is shown in Fig.~\ref{fig:weight}. \par
Fig.~\ref{fig:opt_state} shows the optimization state $\lambda$ defined in \eqref{eqn:dH dt for whole}. The run time threshold for toggling $\lambda$ is set to 100 time steps and robots send convergence signals if $\norm{u_{p,i}} \leq 0.001$ or $|u_{w,i}| \leq 0.01$. When the system converges, all robots send convergence signals and $\lambda$ will be kept toggling. \par
Fig.~\ref{fig:obj_val log} shows the value of the objective function $H$. The change of capability happens at time step 3473 and the removal of robot 1 happens at time step 5148. Consequently, there are sudden jumps on $H$ at those instants. Otherwise, it shows that the value of $H$ is always non-increasing. At the end of the simulation, the value of $H$ is 1431, and the errors between the final allocated area and optimal area are [1.4956 1.464 0.61679 1.4547 1.153 1.1999 -7.384] in square unit.\par

\section{Conclusion and Future Work} \label{section:conclusion}

\addtolength{\textheight}{-9.2cm}   

\subsection{Conclusion}
A coverage control system for a team of robots with heterogeneous patrolling capabilities is presented in this paper. We aim to allocate different portions of a bounded environment to different robots according to their capabilities. The objective function is designed as a function of two blocks of variables, the generators' positions and weights. The block coordinate descent method is used to optimize two blocks of variables alternately. The gradient descent method is used for optimizing each individual block of variables and the gradient of the objective function can be computed in a distributed manner. A centralized machine is used for synchronizing the optimization state and operational state of all robots. It will also update the normalized capabilities of robots, which allows the system to deal with dynamic capabilities and team size. Simulations are conducted to illustrate the results of the coverage control system. \par

\subsection{Future Work}
In this paper, the environment is assumed to be equally important at any points. It might be worthwhile for introducing a density function over the environment to indicate the weights of different points. For example, the patrol for accidents will be more efficient if the human density is considered and placed more robots in the high-density region. Besides, we assumed that the ``robot-to-robot" and ``robot-to-synchronizing machine" communications are established perfectly. Flaws in communication might disturb the synchronization of the system and affect performance. In addition, the robots' sensing ranges and the shapes of the allocated cells are not considered in the system. We assumed that the robots visit all the points in their cell for each patrolling cycle in this paper. It is not necessary if the sensing ranges of robots are larger than a single point, and hence better shapes of cells might improve the efficiency of the patrol.


\end{document}